\documentclass[letterpaper]{article} 
\usepackage[]{aaai2026}  
\usepackage{times}  
\usepackage{helvet}  
\usepackage{courier}  
\usepackage[hyphens]{url}  
\usepackage{graphicx} 
\usepackage{natbib}  
\usepackage{caption} 
\usepackage{enumitem}
\usepackage{amsmath}
\usepackage{float}           
\usepackage[ruled,vlined,linesnumbered]{algorithm2e} 
\usepackage{multirow}

\usepackage{booktabs}
\usepackage{tabularx}
\usepackage{graphicx}
\usepackage{subcaption}

\usepackage{pdfpages}
\nocopyright

\usepackage{newfloat}
\usepackage{listings}
\DeclareCaptionStyle{ruled}{labelfont=normalfont,labelsep=colon,strut=off} 
\floatstyle{ruled}
\newfloat{listing}{tb}{lst}{}
\floatname{listing}{Listing}
\title{StruSR: Structure-Aware Symbolic Regression with Physics-Informed Taylor Guidance}
\author{
    Yunpeng Gong\textsuperscript{\rm 1},
    Sihan Lan\textsuperscript{\rm 1}, Can Yang\textsuperscript{\rm 1}, Kunpeng Xu\textsuperscript{\rm 2}, Min Jiang\textsuperscript{\rm 1}
}
\affiliations{
    \textsuperscript{\rm 1}School of Informatics, Xiamen University, Xiamen, China\\
    \textsuperscript{\rm 2}Université de Sherbrooke, Québec, Canada

}

\usepackage{bibentry}

\begin{document}

\maketitle

\begin{abstract}
	Symbolic regression aims to find interpretable analytical expressions by searching over mathematical formula spaces to capture underlying system behavior, particularly in scientific modeling governed by physical laws. However, traditional methods lack mechanisms for extracting structured physical priors from time series observations, making it difficult to capture symbolic expressions that reflect the system's global behavior. In this work, we propose a structure-aware symbolic regression framework—StruSR—that leverages trained Physics-Informed Neural Networks (PINNs) to extract locally structured physical priors from time series data. By performing local Taylor expansions on the outputs of the trained PINN, we obtain derivative-based structural information to guide symbolic expression evolution. To assess the importance of expression components, we introduce a masking-based attribution mechanism that quantifies each subtree’s contribution to structural alignment and physical residual reduction. These sensitivity scores steer mutation and crossover operations within genetic programming, preserving substructures with high physical or structural significance while selectively modifying less informative components. A hybrid fitness function jointly minimizes physics residuals and Taylor coefficient mismatch, ensuring consistency with both the governing equations and the local analytical behavior encoded by the PINN. Experiments on benchmark PDE systems demonstrate that StruSR improves convergence speed, structural fidelity, and expression interpretability compared to conventional baselines, offering a principled paradigm for physics-grounded symbolic discovery.
\end{abstract}

	
\begin{figure}[ht]
	\centering
	\includegraphics[width=0.48\textwidth]{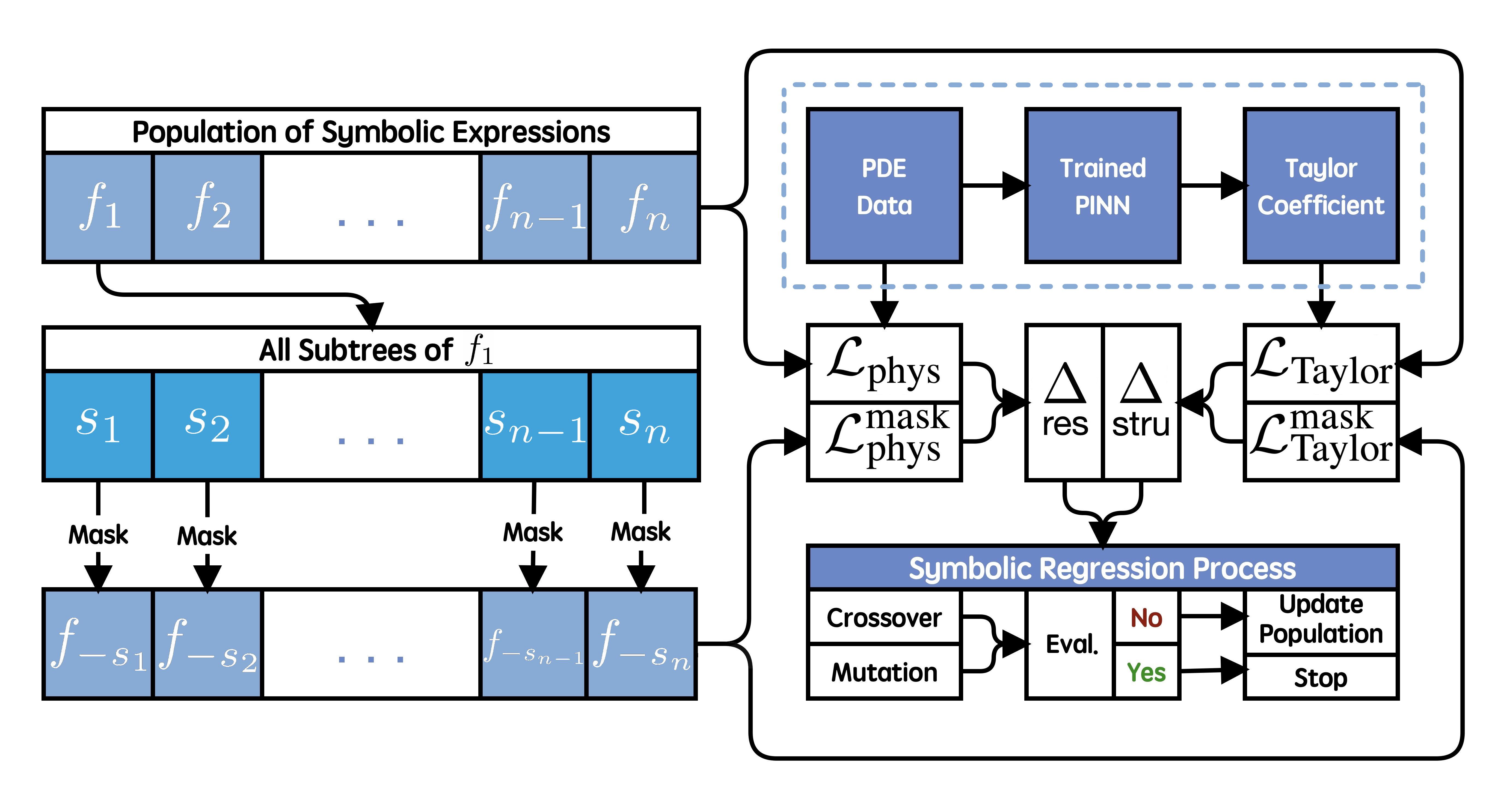}
	\caption{
		Overview of the proposed Taylor-structure-guided symbolic regression framework. From the left, a population of symbolic expressions is maintained and each expression is decomposed into subtrees. Masking individual subtrees  allows evaluating their contributions to physics residual loss ($\mathcal{L}_{phys}$) and Taylor-based structural loss ($\mathcal{L}_{Taylor}$), the latter derived from local Taylor coefficients extracted from a Physics-Informed Neural Network (PINN) trained on PDE data. Sensitivity scores computed from these loss variations  guide crossover and mutation in genetic programming, preserving structurally and physically important subexpressions. The symbolic regression process iterates with population updates until convergence or stopping criteria are met.
	}
	\label{pip}
\end{figure}

\section{Introduction}
In physical field simulations, partial differential equations (PDEs) are employed to model the spatiotemporal evolution of physical quantities, where the time series \cite{wen2020time,luout,xu2024kan4drift,jin2024position,xu2024kolmogorov,zhou2022fedformer,zhang2019deep,qin2017dual} at specific spatial locations reflect the system's dynamic characteristics; symbolic regression, by searching for the underlying analytical structure of the governing PDEs, enables interpretable and generalizable modeling of complex physical systems. Symbolic regression (SR) aims to discover closed-form mathematical expressions that describe the underlying relationships in data. With the rapid advancement of deep learning \cite{1,2,3,4,5,6,7,8,9,10,11,12}, numerous new methods have emerged for solving partial differential equations. Unlike black-box models such as deep neural networks, SR offers human-interpretable outputs, making it particularly valuable in scientific modeling and systems governed by physical laws. In recent years, symbolic regression has seen renewed interest through integration with neural architectures and evolutionary algorithms, yet a persistent challenge remains: how to guide the search toward expressions that are not only accurate but also structurally aligned with physical principles.

Most existing SR methods focus on data fitting accuracy, often overlooking whether the discovered expression reflects the correct physical structure. This shortcoming becomes especially pronounced in the context of discovering governing equations or partial differential equations (PDEs), where expressions must satisfy physical constraints and exhibit meaningful derivative behavior. While PINNs \cite{PINN} offer a promising way to approximate solutions to differential equations by embedding physics into training objectives, their outputs are inherently non-symbolic and lack direct interpretability.

To address this gap, we propose a novel framework that leverages the structural insights encoded in trained PINNs to guide the symbolic regression process. The key idea of our approach is to leverage the local Taylor expansions of a trained PINN solution as structural “ground truth“ for guiding symbolic expression discovery. By comparing the Taylor coefficients of candidate expressions against those extracted from the PINN, we can quantitatively assess structural similarity in a manner that is both principled and robust. This allows us to inject domain knowledge—derived from a continuous, differentiable, and physically meaningful PINN representation—into the symbolic search process in a principled way.
	
From a deeper perspective, a trained PINN should not be regarded merely as a numerical predictor for PDE solutions—it acts as a global encoder of physical structure. Through minimizing residuals of differential operators across the entire domain, PINNs embed physical consistency into the learned parameter space. Although the analytic solution remains hidden within the black-box network, its local structure can be made accessible via automatic differentiation. This gives rise to local Taylor expansions of the form:
\begin{equation}
T_{\text{PINN}}(x; x_0) = \sum_{k=0}^K \frac{u^{(k)}(x_0)}{k!}(x - x_0)^k,
\end{equation}
which serve as a physically informed, differentiable surrogate for the unknown ground-truth expression. Although the local Taylor expansions do not equate to the analytic solution, they provide a locally valid proxy that respects both the governing PDE and the global data distribution. From this view, the Taylor expansion becomes a conceptual and computational bridge—connecting the numeric expressivity of PINNs with the symbolic abstraction of SR.

In contrast to direct derivative estimation from discrete data, which is highly sensitive to noise and lacks structural context, Taylor coefficients extracted from PINNs are globally consistent and physically grounded. This makes them especially suitable as guiding signals during symbolic evolution, where preserving critical structural components is essential for interpretability and convergence.

Building on this insight, we introduce a masking-based attribution mechanism that evaluates the contribution of each expression subtree to both the structural and physical fidelity of the model. These sensitivity scores are then used to inform the mutation and crossover strategies within a genetic programming (GP) framework. Substructures that align closely with the PINN-derived Taylor pattern are preserved, while those with low structural or physical importance are prioritized for modification. In addition, a hybrid fitness function combines physical residual loss with Taylor-based structural loss to jointly optimize both physical consistency and symbolic alignment.

Our approach transforms symbolic regression from a purely heuristic or data-driven procedure into a physics-guided, structurally-informed discovery process. Experiments on benchmark PDEs demonstrate that our method improves convergence speed, reduces overfitting, and produces symbolic expressions that are more consistent with the governing equations. Moreover, our framework offers a flexible plug-in mechanism for incorporating PINN knowledge into symbolic search, opening the door to new forms of hybrid neuro-symbolic modeling.

In summary, this work makes the following contributions:

\begin{itemize}
	\item Proposes a novel framework that leverages local Taylor expansions extracted from Physics-Informed Neural Networks (PINNs) as structured, derivative-based priors to guide symbolic expression discovery through both initialization and loss alignment.
	
	\item Introduces a dual-sensitivity attribution mechanism that quantifies the physical and structural importance of symbolic subtrees, enabling fine-grained evaluation and evolution control.
	
	\item Integrates sensitivity-aware mutation and crossover strategies into genetic programming, along with a composite objective function that jointly optimizes for physical consistency and structural fidelity.
\end{itemize}

This fusion of PINN-informed priors and symbolic regression paves the way for more interpretable, physically meaningful machine-discovered models in scientific computing and data-driven equation discovery.

\section{Related Work}
Partial Differential Equations (PDEs) are fundamental tools for modeling continuous processes in natural and engineering systems. Existing methods for solving PDEs can be broadly categorized into two families: numerical discretization approaches, such as Finite Difference Methods (FDM)\cite{FDM2020,FDM2021,FDM2021U} and Finite Element Methods (FEM)\cite{FEM2020D,FEM2021S,FEM2022P}, which aim to approximate numerical solutions with high precision; and analytical modeling approaches, particularly Symbolic Regression (SR), which seeks to discover explicit mathematical expressions that reveal the underlying physical laws, offering high interpretability and structural transferability\cite{SR2020U,SR2021M,SR2021P}

Traditional symbolic regression methods are mostly based on Genetic Programming (GP), which evolves expression trees to fit data\cite{GPSR}. Representative works such as Eureqa~\cite{Eureqa}, GPTree~\cite{GPTree} have demonstrated strong expressiveness in physical modeling and equation discovery tasks. However, these methods typically lack structural awareness and do not incorporate structural attribution mechanisms or modeling priors, often resulting in overly complex, physically implausible, or poorly convergent expressions. In recent years, approaches like RAG-SR \cite{zhang2025rag} NetGP~\cite{cao2025netgp}, HD-TLGP~\cite{cao2024interpretable}, SP-GPSR~\cite{cao2023genetic} and PhySO \cite{tenachi2023deep} have attempted to combine gradient-based optimization with neural guidance to enhance search efficiency. Nonetheless, they largely remain within a “structure-blind” search paradigm and fail to incorporate physical constraints effectively.

In parallel, neural networks have made significant progress in the numerical solution of PDEs. Physics-Informed Neural Networks (PINNs)\cite{PINN}, in particular, embed PDE residuals into the loss function, enabling the network to learn continuous solutions under unsupervised or weakly supervised settings. A series of follow-up works, including DeepRitz\cite{DeepRITZ}, Galerkin PINNs\cite{GalerkinPINN}, and XPINN\cite{XPINN}, have improved the modeling of solution accuracy, boundary conditions, and generalization. Despite their success, these methods remain fundamentally black-box approximators and cannot generate explicit, interpretable symbolic expressions or reusable structural information.

In recent years, several studies have attempted to bridge PDE solvers (especially PINNs) with symbolic modeling, aiming toward more interpretable and transparent analytical modeling pipelines\cite{PINNSR}. A representative direction is to use PINNs to generate training data and then perform symbolic regression using differentiable program architectures, as in~\cite{PINNSR2}. Another approach embeds symbolic branches (e.g., $\sin$, $\exp$) directly into the network architecture, trained under physics constraints—e.g., Physics-Informed Symbolic Networks (PISN)~\cite{PISN}. Alternatively, some methods fit hidden differential components inside PINNs and then apply symbolic regression (e.g., AI-Feynman) to extract interpretable formulas~\cite{SR2020U}. These methods demonstrate the potential of combining symbolic expressiveness with PDE solution accuracy, yet most remain focused on final expression output or symbolic embedding, without structural attribution or controllable symbolic evolution.

In the field of symbolic regression, TaylorGP~\cite{TaylorGP} is a representative method that incorporates structural guidance mechanisms. It performs Taylor expansions at multiple sampling points of the target function to extract local structural attributes such as variable separability, parity, and monotonicity. These structural cues are then used to guide the initialization of symbolic expressions and constrain evolutionary operations such as mutation and crossover. The method shows preliminary success in improving structural control, demonstrating the positive role of structural information in symbolic modeling. However, TaylorGP lacks a dynamic attribution mechanism and global consistency modeling capability, and it is not suitable for PDE settings where only equation-based constraints are available.

To address these limitations, we propose StruSR—a structurally guided symbolic modeling framework that, for the first time, leverages PDE numerical solvers (PINNs) as structure information extractors rather than direct solvers. Trained over global data and physical constraints, PINNs produce smooth solutions that exhibit global consistency. The Taylor expansion derived from such solutions, though local in form, reflects structural characteristics under global constraints, providing a physically consistent and globally coherent structural prior. Furthermore, StruSR introduces a structure-sensitive attribution mechanism based on masked gradients to identify key substructures and guide symbolic mutation and crossover at fine granularity. This enables a transition from expression-level to structure-level guidance. StruSR effectively integrates the advantages of numerical and symbolic modeling, demonstrating strong performance in PDE-driven symbolic expression recovery tasks.

\begin{algorithm}[ht]\scriptsize
	\caption{Structure-Guided Symbolic Regression via PINN Taylor Expansion}
	\label{alg:structure_gp}
	\KwIn{\\
		 Trained PINN model $u(x)$,
		 Collocation points $\{x_i\}_{i=1}^N$,
		 Differential operator $\mathcal{N}[\cdot]$,
		 Taylor order $K=5$,
		 Population size $P$, mutation rate $p_{\text{mut}}$, crossover rate $p_{\text{cross}}$,
		 Structural weight $\lambda$, sensitivity balance $\beta$.
	}
	\KwOut{\\
		\quad Best symbolic expression $f^*(x)$
	}
	\BlankLine
	Initialize population $\mathcal{G} = \{f_1, f_2, \dots, f_P\}$ with random symbolic expressions\;
	
	Compute PINN Taylor expansions $\{T_{(u)}(x; x_i)\}_{i=1}^N$ at collocation points\;

	\While{termination criteria not met}{
		\ForEach{expression $f \in \mathcal{G}$}{
			Compute structural loss $\mathcal{L}_{\text{Taylor}}(f; x_i)$\;
			Compute physics loss $\mathcal{L}_{\text{phys}}(f) = \frac{1}{N} \sum_{i=1}^{N} (\mathcal{N}[f](x_i))^2$\;
			Evaluate fitness: $\mathcal{F}(f) = \mathcal{L}_{\text{phys}}(f) + \lambda \cdot \mathcal{L}_{\text{Taylor}}(f)$\;
			
			\ForEach{subtree $s_j$ of $f$}{
				Construct masked expression $f_{-s_j}(x)$ by replacing $s_j$ with constant 1\;
				Compute sensitivity scores:\;
				$\quad \Delta_j^{\text{struct}} = \mathcal{L}_{\text{Taylor}}(f_{-s_j}) - \mathcal{L}_{\text{Taylor}}(f)$\;
				$\quad \Delta_j^{\text{res}} = \text{MSE}(\mathcal{N}[f_{-s_j}]) - \text{MSE}(\mathcal{N}[f])$\;
				$\quad \Delta_j^{\text{total}} = \beta \cdot \Delta_j^{\text{res}} + (1 - \beta) \cdot \Delta_j^{\text{struct}}$\;
			}
		}
		
		Select parents from $\mathcal{G}$ based on fitness\;
		
		\ForEach{parent pair $(f_A, f_B)$}{
			\If{rand() $< p_{\text{cross}}$}{
				Sample $s_A$ from $f_A$ using a sensitivity-based probability distribution\;
				Sample $s_B$ from $f_B$\;
				Swap $s_A$ and $s_B$ to create offspring\;
			}
		}
		
		\ForEach{offspring $f$}{
			\If{rand() $< p_{\text{mut}}$}{
				Sample $s_j$ using a sensitivity-based probability distribution\;
				Replace $s_j$ with a random subtree from the symbol library\;
			}
		}
		
		Update population $\mathcal{G}$\;

	}
	
	\Return{expression $f^*(x) = \arg\min_f \mathcal{F}(f)$}
\end{algorithm}

\paragraph{Algorithm Description.}
This algorithm aims to integrate the local Taylor expansion structure derived from a Physics-Informed Neural Network (PINN) into the genetic programming (GP) process for symbolic regression, thereby enhancing the physical consistency and structural interpretability of the generated expressions. The core idea is to introduce a dual sensitivity evaluation mechanism—based on both structural alignment and residual fidelity—during the evolution process, enabling fine-grained guidance over substructures in candidate expressions.

As illustrated in Fig.~\ref{pip}, a pre-trained PINN model is used to extract $K$-order Taylor expansions (default $K=5$) at multiple anchor points $\{x_0^{(i)}\}$ sampled from the domain. These expansions encode the local derivative behavior of the target differential equation around each point, collectively providing a structurally consistent and physically grounded reference for guiding symbolic expression discovery.

The evolution begins with a population of randomly initialized symbolic expressions. For each candidate expression, two key losses are computed: (1) the \textit{physics residual loss} $\mathcal{L}_{\text{phys}}$, which quantifies the mean squared residual when the expression is substituted into the target differential operator $\mathcal{N}[\cdot]$ across all collocation points; and (2) the \textit{structure loss} $\mathcal{L}_{\text{Taylor}}$, measuring the squared difference between the Taylor coefficients of the expression and the PINN expansion. These two metrics are jointly incorporated into a hybrid fitness function $\mathcal{F}(f)$ that reflects both physical correctness and structural alignment.

To assess the local significance of each subcomponent, we introduce a masking attribution mechanism. For each subtree $s_j$ in a symbolic expression $f(x)$, we temporarily mask it by replacing it with a neutral constant (e.g., 1), forming a masked expression $f_{-s_j}(x)$. The changes in structure loss and physics residual due to this masking are recorded as \textit{structural sensitivity} $\Delta_j^{\text{struct}}$ and \textit{residual sensitivity} $\Delta_j^{\text{res}}$, respectively. A total sensitivity score $\Delta_j^{\text{total}}$ is computed via weighted summation controlled by hyperparameter $\beta$, reflecting the joint importance of subtree $s_j$ with respect to both structural and physical objectives.

During genetic operations, sensitivity scores define a probability distribution that guides both crossover and mutation, balancing structural preservation with search diversity. 
For crossover, each parent pair $(f_A, f_B)$ samples a subtree $s_A$ from $f_A$ using this sensitivity-based distribution, giving higher selection probability to less critical regions, and swaps it with a subtree $s_B$ from $f_B$. 
This reduces the risk of disrupting key structures while maintaining exploration. 
For mutation, a subtree $s_j$ is sampled from the same sensitivity-based distribution and replaced with a randomly generated subtree from the symbol library, allowing mutations to occur mainly in non-critical parts while preserving structurally important components.

The evolutionary process repeats until convergence or other termination criteria are met. 
At each generation, expressions are evaluated by a hybrid fitness function that balances physics residual fidelity and structural alignment. 
The best-performing expression $f^*(x)$ is finally selected as the solution.

\begin{table*}[t]\tiny
	\centering
	\caption{Summary of tested differential equations with their deterministic conditions.}
	\renewcommand{\arraystretch}{1}
	\begin{tabular}{l>{\centering\arraybackslash}m{7.2cm}>{\centering\arraybackslash}m{7.2cm}}
		\toprule
		\textbf{Name} & \textbf{PDE} & \textbf{Deterministic Conditions} \\
		
		\midrule
		Advection &
		$\dfrac{\partial \phi}{\partial t} + u \dfrac{\partial \phi}{\partial x} = 0$ &
		$u(x, 0) = \sin(x),\quad x \in [0,1],\; t \in [0,1]$ \\
		
		\midrule
		Diffusion &
		$\dfrac{\partial u}{\partial t} = \dfrac{\partial^2 u}{\partial x^2} - e^{-t} \sin(\pi x)(1 - \pi^2)$ &
		\begin{tabular}{@{}c@{}}
			$u(x, 0) = \sin(\pi x),\quad x \in [-1,1]$ \\
			$u(-1, t) = u(1, t) = 0,\quad t \in [0,1]$
		\end{tabular} \\
		
		\midrule
		Poisson2D &
		$\frac{\partial^2 u}{\partial x_1^2} + \frac{\partial^2 u}{\partial x_2^2} = 30x_1^2 - 7.8x_1 + 1$ &
		$u(\mathbf{x}) = 2.5x_1^4 - 1.3x_1^3 + 0.5x_2^2 - 1.7x_2,\quad \mathbf{x} \in \partial[-1,1]^2$ \\
		
		\midrule
		Poisson3D &
		$\frac{\partial^2 u}{\partial x_1^2} + \frac{\partial^2 u}{\partial x_2^2} + \frac{\partial^2 u}{\partial x_3^2} = 30x_1^2 - 7.8x_2 + 1$ &
		$u(\mathbf{x}) = 2.5x_1^4 - 1.3x_2^3 + 0.5x_3^2,\quad \mathbf{x} \in \partial[-1,1]^3$ \\
		
		\midrule
		Heat2D &
		$\frac{\partial u}{\partial t} - \left( \frac{\partial^2 u}{\partial x_1^2} + \frac{\partial^2 u}{\partial x_2^2} \right) = -30x_1^2 + 7.8x_2 + t$ &
		\begin{tabular}{@{}c@{}}
			$u(\mathbf{x}, t) = 2.5x_1^4 - 1.3x_2^3 + 0.5t^2,\quad \mathbf{x} \in \partial[-1,1]^2,\; t \in [0,1]$ \\
			$u(\mathbf{x}, 0) = 2.5x_1^4 - 1.3x_2^3,\quad \mathbf{x} \in [-1,1]^2$
		\end{tabular} \\
		
		\midrule
		Heat3D &
		$\frac{\partial u}{\partial t} - \left( \frac{\partial^2 u}{\partial x_1^2} + \frac{\partial^2 u}{\partial x_2^2} + \frac{\partial^2 u}{\partial x_3^2} \right) = -30x_1^2 + 7.8x_2 - 2.7$ &
		\begin{tabular}{@{}c@{}}
			$u(\mathbf{x}, t) = 2.5x_1^4 - 1.3x_2^3 + 0.5x_3^2 - 1.7t,\quad \mathbf{x} \in \partial[-1,1]^3,\; t \in [0,1]$ \\
			$u(\mathbf{x}, 0) = 2.5x_1^4 - 1.3x_2^3 + 0.5x_3^2,\quad \mathbf{x} \in [-1,1]^3$
		\end{tabular} \\
		
		\midrule
		Wave2D &
		$\frac{\partial^2 u}{\partial t^2} - \left( \frac{\partial^2 u}{\partial x_1^2} + \frac{\partial^2 u}{\partial x_2^2} \right) = -u^3 + (0.25 - 4x_1^2)$ &
		\begin{tabular}{@{}c@{}}
			$u(\mathbf{x}, 0) = \exp(x_1^2)\sin(x_2),\quad \mathbf{x} \in [-1,1]^2$ \\
			$u(\mathbf{x}, t) = \exp(x_1^2)\sin(x_2)e^{-0.5t},\quad \mathbf{x} \in \partial[-1,1]^2,\; t \in [0,1]$
		\end{tabular} \\
		
		\midrule
		Wave3D &
		$\frac{\partial^2 u}{\partial t^2} - \left( \frac{\partial^2 u}{\partial x_1^2} + \frac{\partial^2 u}{\partial x_2^2} + \frac{\partial^2 u}{\partial x_3^2} \right) = u^2 - \left( 4x_1^2 + 4x_3^2 + 2.75 \right)$ &
		\begin{tabular}{@{}c@{}}
			$u(\mathbf{x}, 0) = \exp(x_1^2 + x_3^2)\cos(x_2),\quad \mathbf{x} \in [-1,1]^3$ \\
			$u(\mathbf{x}, t) = \exp(x_1^2 + x_3^2)\cos(x_2)e^{-0.5t},\quad \mathbf{x} \in \partial[-1,1]^3,\; t \in [0,1]$
		\end{tabular} \\
		
		\bottomrule
	\end{tabular}
	\label{tab:pde_summary}
\end{table*}

\begin{table*}[t]\small
	\centering
	\caption{Mean absolute error (MAE) comparison between StruSR and representative symbolic regression baselines on recovering closed-form solutions for a set of PDE benchmarks. Each entry reports the average test MAE and its standard deviation over 10 independent runs. }
	\begin{tabular}{lccccc}
		\toprule
		\textbf{Name} & 
		\textbf{StruSR (Ours)} & 
		\textbf{RAG-SR} &
		\textbf{NetGP} & 
		\textbf{HD-TLGP} & 
		\textbf{PhySO} \\
		\midrule
		
		Advection & 1.21E-13 $\pm$ 3.57E-15 
		& 2.03E-13 $\pm$ 4.22E-14 
		& \textbf{9.59E-14 $\pm$ 1.27E-14 }
		& 7.91E-11 $\pm$ 6.89E-12 
		& 3.18E-10 $\pm$ 1.10E-10 \\
		
		Diffusion & \textbf{6.38E-6 $\pm$ 1.33E-6 }
		& 8.91E-6 $\pm$ 1.18E-6 
		& 1.79E-5 $\pm$ 2.12E-5 
		& 2.23E-5 $\pm$ 3.97E-5 
		& 3.41E-3 $\pm$ 7.59E-4 \\
		
		Poisson2D & \textbf{7.32E-5 $\pm$ 2.71E-6} 
		& 8.03E-5 $\pm$ 3.21E-6 
		& 5.69E-1 $\pm$ 7.94E-2 
		& 1.87E-2 $\pm$ 2.40E-3 
		& 7.13E+0 $\pm$ 1.43E+0 \\
		
		Poisson3D & \textbf{5.62E-3 $\pm$ 1.41E-3} 
		& 5.77E-3 $\pm$ 1.52E-3 
		& 2.67E+0 $\pm$ 1.08E-1 
		& 1.85E-1 $\pm$ 1.69E-2 
		& 6.11E+0 $\pm$ 1.48E+0 \\
		
		Heat2D    & \textbf{1.53E-6 $\pm$ 7.81E-7}
		& 1.61E-6 $\pm$ 8.12E-7
		& 3.16E+0 $\pm$ 1.32E-1
		& 2.21E-2 $\pm$ 1.34E-2 
		& 1.18E+1 $\pm$ 1.21E+0 \\
		
		Heat3D    & 1.09E-4 $\pm$ 1.29E-5
		& \textbf{9.27E-5 $\pm$ 1.42E-5}
		& 2.74E+0 $\pm$ 3.32E-1
		& 8.71E-2 $\pm$ 2.91E-3 
		& 1.13E+1 $\pm$ 1.69E+0 \\
		
		Wave2D    & \textbf{4.33E-5 $\pm$ 2.41E-6}
		& 4.79E-5 $\pm$ 2.89E-6
		& 9.47E+0 $\pm$ 2.19E-1
		& 4.32E-4 $\pm$ 3.57E-4 
		& 2.01E+0 $\pm$ 1.39E-1 \\
		
		Wave3D    & \textbf{7.09E-6 $\pm$ 1.64E-6}
		& 8.92E-6 $\pm$ 1.52E-6
		& 1.36E+1 $\pm$ 4.03E+0
		& 9.71E-4 $\pm$ 2.35E-5 
		& 1.34E+1 $\pm$ 2.29E+0 \\
		
		\bottomrule
	\end{tabular}
	\label{tab:phy}
\end{table*}

\begin{table*}[ht]\scriptsize
	\centering
	\caption{Comparison of ground-truth analytical solutions and symbolic solutions identified by StruSR on classic PDEs.}
	\label{tab:stru-pde-comparison}
	\renewcommand{\arraystretch}{0.9}
	\begin{tabularx}{\textwidth}{@{}l>{\raggedright\arraybackslash}X>{\raggedright\arraybackslash}X@{}}
		\toprule
		\textbf{Name} & \textbf{Ground Truth $u_{\text{true}}$} & \textbf{Symbolic Solution $\hat{u}$ by StruSR} \\
		\midrule
		
		\textbf{Advection} &
		$ \sin(x_1 - t) $ &
		$ \sin(x_1 - 1.001t) + 0.001\sin(2x_1) $ \\
		
		\addlinespace
		
		\textbf{Diffusion} &
		$ e^{-t} \sin(\pi x) $ &
		$ (1.022 - 0.482 t + 0.124 t^{2}) \sin(3.145 x - 0.005 t) $ \\
		
		\addlinespace
		
		\textbf{Poisson2D} &
		$ x_1^2(2.5x_1^2 - 1.3x_1) + 0.5x_2(x_2 - 1.4) - x_2 $ &
		$ x_1^2(2.500x_1^2 - 1.301x_1) + 0.504x_2(x_2 - 1.396) - x_2 $ \\
		
		\addlinespace
		
		\textbf{Poisson3D} &
		$ 2.5x_1^4 - 1.3x_2^3 + 0.5x_3^2 $ &
		$ 2.500x_1^4 - 1.300x_2^3 + 0.501x_3^2 - 2.158\times 10^{-5} $ \\
		
		\addlinespace
		
		\textbf{Wave2D} &
		$ \exp(x_1^2)\sin(x_2)\, e^{-0.5t} $ &
		$ (1.000 - 0.500t + 0.120t^2)(1.000 + x_1^2)\sin(x_2) $ \\
		
		\addlinespace
		
		\textbf{Wave3D} &
		$ \exp(x_1^2 + x_3^2) \sin(x_2) e^{-0.5t} $ &
		$ \exp(-0.5001t + 1.000x_1^2 + 0.999x_3^2)\sin(x_2) $ \\
		
		\addlinespace
		
		\textbf{Heat2D} &
		$ 0.5t^2 + 2.5x_1^4 - 1.3x_2^3 $ &
		$ 0.500^2 + 2.501x_1^4 - 1.299x_2(x_2 + 0.020)x_2 $ \\
		
		\addlinespace
		
		\textbf{Heat3D} &
		$ -1.7t + 2.5x_1^4 - 1.3x_2^3 + 0.5x_3^2 $ &
		$ -1.701t + 2.500x_1^4 - 1.300x_2^3 + 0.499x_3^2 $ \\
		
		\bottomrule
	\end{tabularx}
	\label{sr-comparison}
\end{table*}

\section{Methodology}

We propose a structure-aware symbolic regression framework that integrates structural priors extracted from Physics-Informed Neural Networks (PINNs) into the evolutionary process of Genetic Programming (GP). At the heart of this approach lies the use of local Taylor expansions derived from PINNs as a source of ground-truth structural guidance. This mechanism promotes the discovery of symbolic expressions that are not only accurate with respect to data but also structurally aligned with underlying physical principles.

Let $u(x)$ denote a trained PINN model that approximates the solution to a target equation. Around a selected reference point $x_0$, we extract a $K$-order Taylor expansion:
\begin{equation}
	T_{(u)}(x; x_0) = \sum_{k=0}^{K} \frac{u^{(k)}(x_0)}{k!} (x - x_0)^k
\end{equation}
Here, $u^{(k)}(x_0)$ denotes the $k$-th derivative of $u(x)$ evaluated at $x_0$, and $k!$ is the factorial normalization. Likewise, a candidate symbolic expression $f(x)$ can be expanded in the same manner. The structure loss is then defined as the mean squared error between the Taylor coefficients of $f(x)$ and $u(x)$:
\begin{equation}
	\mathcal{L}_{\text{Taylor}}(f; x_0) = \sum_{k=0}^{K} \left( \frac{f^{(k)}(x_0)}{k!} - \frac{u^{(k)}(x_0)}{k!} \right)^2
\end{equation}
This loss function provides a robust and interpretable measure of structural alignment. Since Taylor expansions are naturally ordered by monomial degree $(x - x_0)^k$, structural information is encoded directly in the coefficient vector. When a derivative term is theoretically absent in $f(x)$, its corresponding coefficient should ideally be zero. However, due to numerical estimation errors, such coefficients are often small but non-zero. We set the Taylor expansion order $K=5$ to balance expressiveness and numerical stability: lower orders may underfit local structure, while higher orders are prone to derivative noise and overfitting.

To identify which subtrees in a symbolic expression contribute most to the structure and data fidelity, we introduce a \textit{masking attribution} mechanism. For each subtree $s_j$ in a candidate expression $f(x)$, we construct a masked version $f_{-s_j}(x)$ by replacing $s_j$ with a neutral constant (e.g., 1). We then evaluate two forms of sensitivity:

\textbf{Structural sensitivity} quantifies the change in structure loss after masking:

\begin{equation}
	\Delta_j^{\text{struct}} = \mathcal{L}_{\text{Taylor}}(f_{-s_j}; x_i) - \mathcal{L}_{\text{Taylor}}(f; x_i).
\end{equation}

We define the Taylor structure loss $\mathcal{L}_{\text{Taylor}}$ at point $x_i$ as:
\begin{equation}
	\mathcal{L}_{\text{Taylor}}(f; x_i) = 
	\sum_{k=0}^K \Bigg(\frac{f^{(k)}(x_i)}{k!} - \frac{u^{(k)}(x_i)}{k!}\Bigg)^2,
\end{equation}

where $u(\cdot)$ is obtained from the trained PINN, and $f(\cdot)$ is obtained from the candidate symbolic expression.

\textbf{Residual sensitivity} quantifies the change in physical residual error, defined by the violation of the governing differential equation when substituting the symbolic expression into the physics operator $\mathcal{N}[\cdot]$:

\begin{equation}
	\Delta_j^{\text{res}} = \text{MSE}(\mathcal{N}[f_{-s_j}](x)) - \text{MSE}(\mathcal{N}[f](x))
\end{equation}

Here, $\mathcal{N}[\cdot]$ represents the differential operator associated with the target PDE, and the residual is evaluated point-wise over a collocation set $x$. The mean squared error measures how well the expression satisfies the physical constraint. We integrate both sensitivities into a single scalar score:

\begin{equation}
	\Delta_j^{\text{total}} = \beta \cdot \Delta_j^{\text{res}} + (1 - \beta) \cdot \Delta_j^{\text{struct}}
\end{equation}

The balancing parameter $\beta \in [0, 1]$ controls the trade-off between residual accuracy and structural fidelity. Subtrees with low $\Delta_j^{\text{total}}$ are deemed structurally and functionally unimportant, and thus are prioritized for modification during evolution.

In crossover, we adopt a sensitivity-guided strategy that prioritizes protecting beneficial symbolic patterns while maintaining diversity in the search. 
Instead of deterministically selecting the lowest-sensitivity subtrees for swapping, we construct a softmax distribution over all subtrees based on $\Delta_j^{\text{total}}$. 
This biases the selection toward less sensitive regions of each parent, yet assigns a small probability to higher-sensitivity subtrees, preventing the search from repeatedly modifying the same locations and improving overall exploration. 
This approach preserves structural alignments inherited from PINN guidance while reducing the risk of premature convergence, ultimately enhancing the interpretability of the evolved expressions.

Similarly, for mutation, a candidate expression is mutated with probability $p_{\text{mut}}$. 
When triggered, a subtree $s_j$ is sampled from the same softmax distribution, favoring low-sensitivity components but allowing occasional changes in other regions to ensure population diversity. 
The selected subtree is then replaced with a randomly generated one from the symbol library, introducing variability while largely preserving structurally critical components.

Finally, to balance physical fidelity and symbolic expressiveness, we define a hybrid fitness function that jointly penalizes physics violations and structural mismatch:

\begin{equation}
	\mathcal{F}(f) = \mathcal{L}_{\text{phys}}(f) + \mathcal{L}_{\text{Taylor}}(f)
\end{equation}

Here, $\mathcal{L}_{\text{phys}}(f)$ denotes the physics residual loss, computed as the mean squared deviation of $f$ when substituted into the governing differential equation operator $\mathcal{N}[\cdot]$, i.e.,

\begin{equation}
\mathcal{L}_{\text{phys}}(f) = \frac{1}{N} \sum_{i=1}^{N} \left( \mathcal{N}[f](x_i) \right)^2
\end{equation}

where $\{x_i\}_{i=1}^{N}$ are collocation points and $\mathcal{N}[f]$ represents the target PDE/ODE form. The second term $\mathcal{L}_{\text{Taylor}}(f)$ quantifies the structural mismatch between the Taylor expansion of the candidate expression and that of the reference PINN at selected anchor points. Candidate expressions are selected based on this composite objective, facilitating symbolic discovery that is both accurate and grounded in physical priors.

In summary, our framework transforms the symbolic regression pipeline by incorporating Taylor-expansion-based structural signals into GP evolution. It promotes the generation of symbolic models that exhibit both empirical accuracy and physics-guided interpretability.

\begin{figure*}[ht]
	\centering
	\includegraphics[width=1\textwidth]{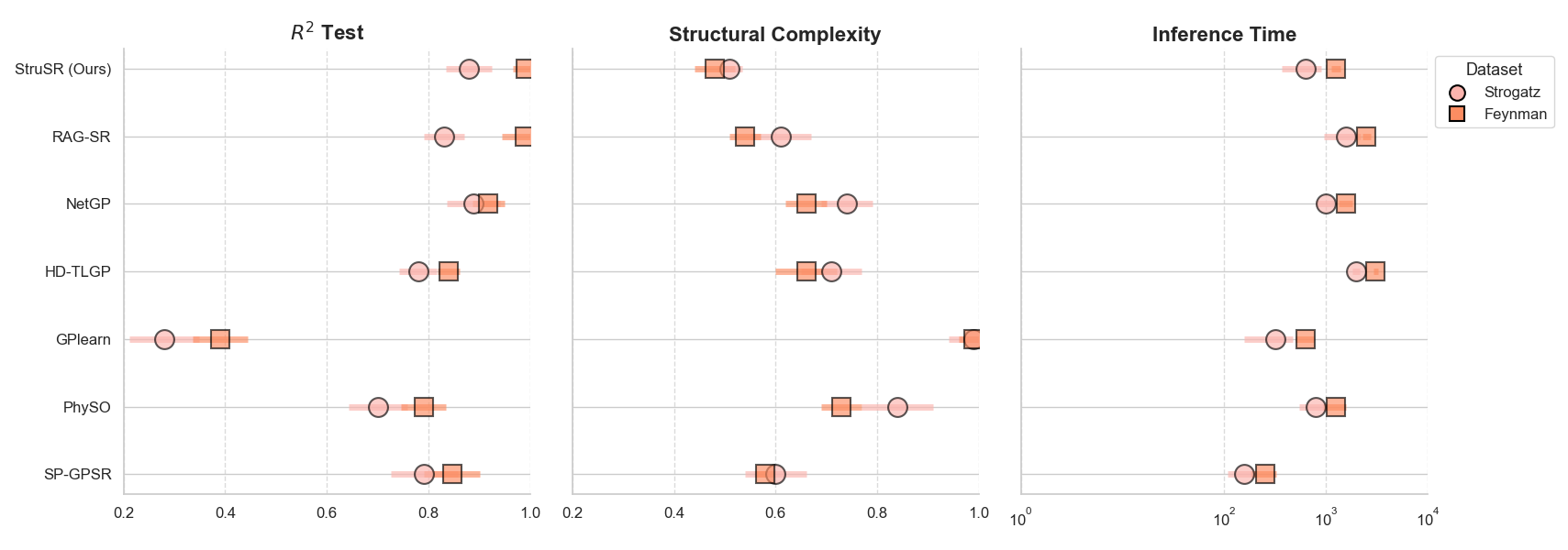}
	\caption{
		Performance comparison of symbolic regression methods across three evaluation dimensions on two benchmark suites (Strogatz and Feynman). Each row corresponds to a specific baseline algorithm, while the circular and square markers represent results on the Strogatz and Feynman datasets, respectively. Subfigure (a) reports the test $R^2$ score (higher is better), indicating predictive accuracy. Subfigure (b) shows the normalized structural complexity (lower is better), reflecting the compactness of the learned expressions. Subfigure (c) presents the inference time (lower is better), measuring computational efficiency.
	}
	\label{5}
\end{figure*}

\begin{figure*}[ht]
	\centering
	\begin{subfigure}[t]{0.48\linewidth}
		\centering
		\includegraphics[width=\linewidth]{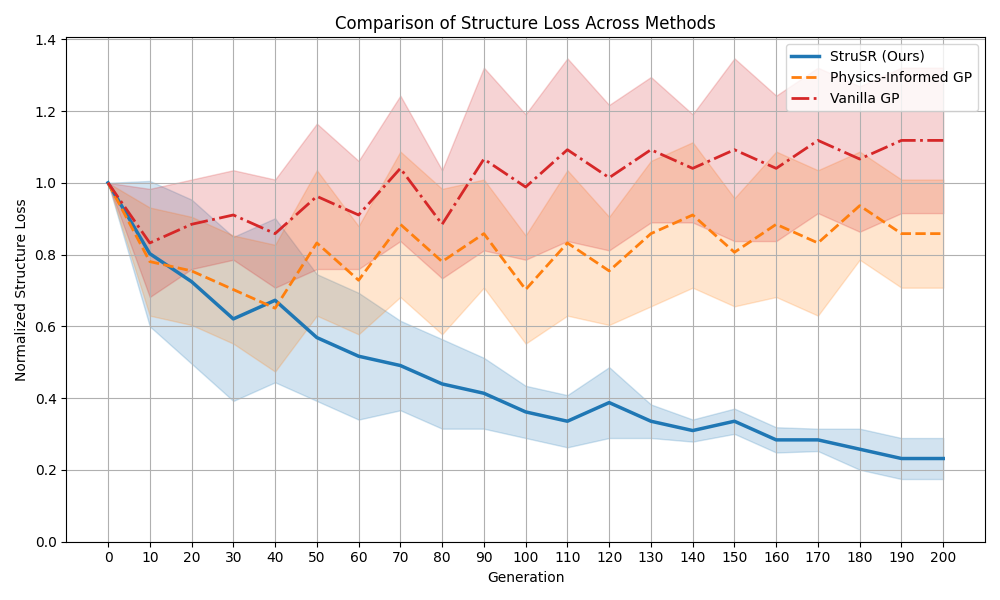}
		\caption{Normalized structure loss across generations.}
		\label{fig:structure_loss}
	\end{subfigure}
	\hfill
	\begin{subfigure}[t]{0.48\linewidth}
		\centering
		\includegraphics[width=\linewidth]{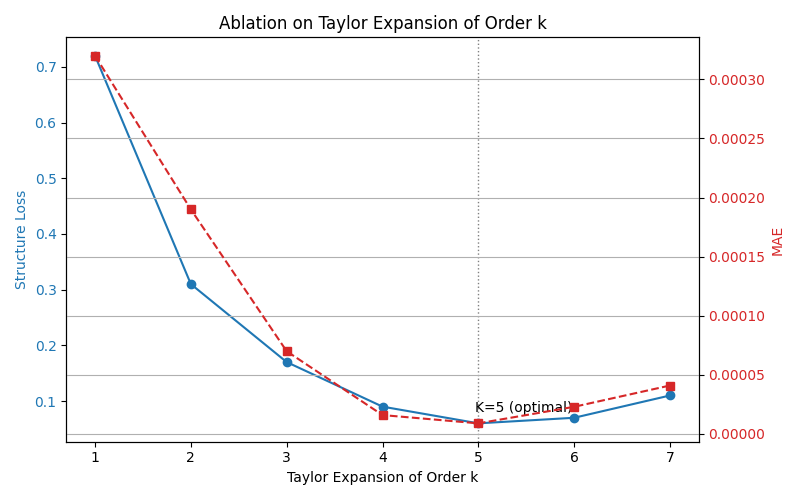}
		\caption{Effect of Taylor expansion order $K$ on structure loss and MAE.}
		\label{fig:taylor_order}
	\end{subfigure}
	\caption{Ablation studies on structural supervision: (a) convergence of normalized structure loss across different methods; (b) impact of Taylor expansion order $K$ on performance.}
	\label{fig:ablation_combined}
\end{figure*}

\section{Experiments}

\subsection{Benchmark Setup}

To comprehensively evaluate the accuracy and interpretability of the proposed method, we construct a PDE benchmark suite consisting of eight representative partial differential equations, covering one, two, and three-dimensional cases. These tasks are designed to reflect various physical structures and boundary conditions. For each equation, the goal is to recover the underlying analytical solution based on given boundary constraints and sparse observations.
Tab.~\ref{tab:pde_summary} summarizes the governing equations along with their deterministic conditions, including initial and boundary constraints.

\subsection{Quantitative Performance Comparison}

We first compare StruSR against representative symbolic regression baselines on a suite of PDE recovery tasks. Competing methods include RAG-SR, NetGP, HD-TLGP, and PhySO. Evaluation is based on the Mean Absolute Error (MAE) computed on the test set. Each result reports the average and standard deviation over 10 independent runs to ensure statistical robustness.

As shown in Tab.~\ref{tab:phy}, StruSR consistently achieves the lowest or second-lowest MAE across most PDE benchmarks. This advantage is particularly evident in structurally complex scenarios such as Poisson2D/3D, Wave2D, and Heat2D. We attribute this superior performance to StruSR’s structure-aware modeling framework, which effectively integrates Taylor priors and guided evolution to align expression generation with the underlying physical structure.

Moreover, StruSR maintains competitive accuracy in 3D problems, demonstrating strong generalization to high-dimensional tasks. While RAG-SR shows promising results on specific cases like Heat3D, it falls short in overall consistency. Other baselines, such as NetGP and PhySO, exhibit higher variance across tasks, indicating limited robustness in structural expression discovery.

\subsection{Analytical Structure Matching}

To assess structural alignment between predicted and ground-truth solutions, we further compare symbolic expressions identified by StruSR with the original closed-form equations. Tab.\ref{tab:stru-pde-comparison} shows that the discovered formulas exhibit strong structural fidelity, often matching the ground truth up to minor coefficient variations or algebraic transformations. These results validate StruSR’s ability to not only approximate numerical behavior but also recover semantically faithful symbolic forms.

\subsection{General Symbolic Modeling Performance}

To assess the general modeling capability of StruSR beyond PDE recovery, we further evaluate it on two widely-used symbolic regression benchmarks: the Strogatz system and the Feynman equation dataset \cite{SR2020U}. Following the standard symbolic regression evaluation protocol~\citep{zhang2025rag}, each dataset is randomly split into 75\% training and 25\% testing subsets. All methods are evaluated based on 10 independent runs. We report three metrics: test $R^2$ score (higher is better), normalized structural complexity (lower is better), and average inference time (lower is better).

Fig.\ref{5}(a) shows that StruSR achieves competitive or leading prediction accuracy on both datasets, demonstrating strong symbolic function approximation capabilities. This performance stems from its structure-aware expression evolution strategy, which allows the model to efficiently explore semantically meaningful subspaces.
In terms of symbolic expression complexity, Fig.\ref{5}(b) evaluates the number of nodes in the output expression trees. All complexity values are normalized to the $[0,1]$ interval for each task, where 1 corresponds to the most complex expression in the set. StruSR consistently produces more compact and interpretable formulas, reflecting the effectiveness of its structural guidance mechanisms in constraining unnecessary redundancy during evolution.
Fig.\ref{5}(c) presents the average inference time per expression on a logarithmic scale. StruSR exhibits relatively low inference latency across most tasks, highlighting its potential for real-time and resource-efficient deployment scenarios.
These results collectively demonstrate that StruSR not only achieves accurate symbolic modeling, but also provides structurally concise and computationally efficient solutions across diverse application domains.

\subsection{Structural Guidance Analysis}

To evaluate the effectiveness of structural guidance in symbolic expression evolution, we conduct two complementary studies: structure loss dynamics across methods and an ablation on Taylor expansion order for prior extraction.

Fig.\ref{fig:ablation_combined}(a) shows normalized structure loss over generations for three methods: Vanilla GP (red), Physics-Informed GP (orange), and our StruSR (blue). StruSR achieves stable and monotonic structure loss reduction, outperforming both baselines. Vanilla GP shows minimal structural improvement, lacking inductive bias. Physics-Informed GP improves early on but fluctuates later due to the absence of explicit structural preservation. Fig.\ref{fig:ablation_combined}(b) evaluates the impact of varying Taylor expansion order \(K\). Both structure loss and prediction MAE are minimized at \(K=5\), indicating an optimal trade-off between local fidelity and robustness. Lower orders (\(K \leq 2\)) lack sufficient derivative information, while higher orders (\(K > 5\)) risk overfitting and noise.
These results validate our dual-level guidance: Taylor-based initialization from a PINN captures globally consistent physical priors, while masking-based attribution enables fine-grained, subtree-level mutation control. Together, they promote stable, interpretable, and physically grounded symbolic modeling.

\section{Conclusion}

We propose StruSR, a structure-aware symbolic regression framework that leverages physics-informed priors derived from Taylor expansions of trained PINNs. By introducing a dual-level structural guidance mechanism—comprising Taylor-based initialization and sensitivity-based attribution—StruSR steers symbolic evolution toward globally consistent, physically meaningful solutions. Extensive experiments on PDE recovery and standard symbolic benchmarks demonstrate that StruSR consistently delivers high predictive accuracy, reduced structural complexity. It reliably discovers semantically faithful expressions with stable and interpretable structural evolution, highlighting the effectiveness of embedding physical inductive biases into symbolic modeling.
This work lays a principled foundation for physics-guided symbolic discovery and opens new directions for interpretable, efficient, and generalizable equation modeling.

\bibliographystyle{unsrt}
\bibliography{aaai2026}

\begin{thebibliography}{45}
\providecommand{\natexlab}[1]{#1}

\bibitem[{Achituve et~al.(2021)Achituve, Navon, Yemini, Chechik, and
  Fetaya}]{GPTree}
Achituve, I.; Navon, A.; Yemini, Y.; Chechik, G.; and Fetaya, E. 2021.
\newblock GP-Tree: A Gaussian process classifier for few-shot incremental
  learning.
\newblock In \emph{International conference on machine learning}, 54--65. PMLR.

\bibitem[{Cao et~al.(2025)Cao, Feng, Jiang, and Tan}]{cao2025netgp}
Cao, L.; Feng, Y.; Jiang, M.; and Tan, K.~C. 2025.
\newblock NetGP: A Hybrid Framework Combining Genetic Programming and Deep
  Reinforcement Learning for PDE Solutions.
\newblock In \emph{2025 IEEE Congress on Evolutionary Computation (CEC)}, 1--8.
  IEEE.

\bibitem[{Cao et~al.(2024)Cao, Liu, Wang, Xu, Ye, Tan, and
  Jiang}]{cao2024interpretable}
Cao, L.; Liu, Y.; Wang, Z.; Xu, D.; Ye, K.; Tan, K.~C.; and Jiang, M. 2024.
\newblock An interpretable approach to the solutions of high-dimensional
  partial differential equations.
\newblock In \emph{Proceedings of the AAAI Conference on Artificial
  Intelligence}, volume~38, 20640--20648.

\bibitem[{Cao et~al.(2023)Cao, Zheng, Ding, Cai, and Jiang}]{cao2023genetic}
Cao, L.; Zheng, Z.; Ding, C.; Cai, J.; and Jiang, M. 2023.
\newblock Genetic programming symbolic regression with simplification-pruning
  operator for solving differential equations.
\newblock In \emph{International Conference on Neural Information Processing},
  287--298. Springer.

\bibitem[{Das et~al.(2025)Das, Bhaumik, De, and Changdar}]{PINNSR}
Das, J.; Bhaumik, B.; De, S.; and Changdar, S. 2025.
\newblock Physics-informed neural network with symbolic regression for deriving
  analytical approximate solutions to nonlinear partial differential equations.
\newblock \emph{Neural Computing and Applications}, 1--36.

\bibitem[{David~M{\"u}zel et~al.(2020)David~M{\"u}zel, Bonhin, Guimar{\~a}es,
  and Guidi}]{FEM2020D}
David~M{\"u}zel, S.; Bonhin, E.~P.; Guimar{\~a}es, N.~M.; and Guidi, E.~S.
  2020.
\newblock Application of the finite element method in the analysis of composite
  materials: A review.
\newblock \emph{Polymers}, 12(4): 818.

\bibitem[{Dub{\v{c}}{\'a}kov{\'a}(2011)}]{Eureqa}
Dub{\v{c}}{\'a}kov{\'a}, R. 2011.
\newblock Eureqa: software review.

\bibitem[{Gong et~al.(2026)Gong, Hou, Shi, DIEP, and Jiang}]{9}
Gong, Y.; Hou, Y.; Shi, J.; DIEP, K.~L.; and Jiang, M. 2026.
\newblock A Theory-Guided Framework for Few-Shot Cross-Modal Sketch Person
  Re-Identification.

\bibitem[{Gong et~al.(2024{\natexlab{a}})Gong, Hou, Wang, Lin, and Jiang}]{3}
Gong, Y.; Hou, Y.; Wang, Z.; Lin, Z.; and Jiang, M. 2024{\natexlab{a}}.
\newblock Adversarial learning for neural PDE solvers with sparse data.
\newblock \emph{arXiv preprint arXiv:2409.02431}.

\bibitem[{Gong et~al.(2023)Gong, Hou, Zhang, and Jiang}]{6}
Gong, Y.; Hou, Y.; Zhang, C.; and Jiang, M. 2023.
\newblock Beyond augmentation: Empowering model robustness under extreme
  capture environments.
\newblock In \emph{IJCNN,2023}. IEEE.

\bibitem[{Gong, Huang, and Chen(2021)}]{2}
Gong, Y.; Huang, L.; and Chen, L. 2021.
\newblock Eliminate deviation with deviation for data augmentation and a
  general multi-modal data learning method.
\newblock \emph{arXiv preprint arXiv:2101.08533}.

\bibitem[{Gong, Huang, and Chen(2022)}]{4}
Gong, Y.; Huang, L.; and Chen, L. 2022.
\newblock Person re-identification method based on color attack and joint
  defence.
\newblock In \emph{CVPR, 2022}, 4313--4322.

\bibitem[{Gong et~al.(2022)Gong, Li, Chen, and Jiang}]{10}
Gong, Y.; Li, J.; Chen, L.; and Jiang, M. 2022.
\newblock Exploring color invariance through image-level ensemble learning.
\newblock \emph{arXiv preprint arXiv:2401.10512}.

\bibitem[{Gong et~al.(2024{\natexlab{b}})Gong, Zeng, Xu, Wang, and Jiang}]{8}
Gong, Y.; Zeng, Q.; Xu, D.; Wang, Z.; and Jiang, M. 2024{\natexlab{b}}.
\newblock Cross-modality attack boosted by gradient-evolutionary multiform
  optimization.
\newblock \emph{arXiv preprint arXiv:2409.17977}.

\bibitem[{Gong et~al.(2024{\natexlab{c}})Gong, Zhang, Hou, Chen, and Jiang}]{5}
Gong, Y.; Zhang, C.; Hou, Y.; Chen, L.; and Jiang, M. 2024{\natexlab{c}}.
\newblock Beyond dropout: Robust convolutional neural networks based on local
  feature masking.
\newblock In \emph{IJCNN,2024}. IEEE.

\bibitem[{Gong et~al.(2024{\natexlab{d}})Gong, Zhong, Qu, Luo, Ji, and
  Jiang}]{1}
Gong, Y.; Zhong, Z.; Qu, Y.; Luo, Z.; Ji, R.; and Jiang, M. 2024{\natexlab{d}}.
\newblock Cross-modality perturbation synergy attack for person
  re-identification.
\newblock \emph{Advances in Neural Information Processing Systems}, 37:
  23352--23377.

\bibitem[{He et~al.(2022)He, Lu, Yang, Luo, and Wang}]{TaylorGP}
He, B.; Lu, Q.; Yang, Q.; Luo, J.; and Wang, Z. 2022.
\newblock Taylor genetic programming for symbolic regression.
\newblock In \emph{Proceedings of the genetic and evolutionary computation
  conference}, 946--954.

\bibitem[{Jagtap and Karniadakis(2020)}]{XPINN}
Jagtap, A.~D.; and Karniadakis, G.~E. 2020.
\newblock Extended physics-informed neural networks (XPINNs): A generalized
  space-time domain decomposition based deep learning framework for nonlinear
  partial differential equations.
\newblock \emph{Communications in Computational Physics}, 28(5).

\bibitem[{Jin et~al.(2024)Jin, Zhang, Chen, Zhang, Liang, Yang, Wang, Pan, and
  Wen}]{jin2024position}
Jin, M.; Zhang, Y.; Chen, W.; Zhang, K.; Liang, Y.; Yang, B.; Wang, J.; Pan,
  S.; and Wen, Q. 2024.
\newblock Position: What can large language models tell us about time series
  analysis.
\newblock In \emph{41st International Conference on Machine Learning}.
  MLResearchPress.

\bibitem[{Lin et~al.(2024{\natexlab{a}})Lin, Wang, Hou, Tang, and Jiang}]{12}
Lin, J.; Wang, Z.; Hou, Y.; Tang, Y.; and Jiang, M. 2024{\natexlab{a}}.
\newblock Phy124: Fast physics-driven 4d content generation from a single
  image.
\newblock \emph{arXiv preprint arXiv:2409.07179}.

\bibitem[{Lin et~al.(2024{\natexlab{b}})Lin, Zhenzhong, Dejun, Shu, Gong, and
  Jiang}]{11}
Lin, J.; Zhenzhong, W.; Dejun, X.; Shu, J.; Gong, Y.; and Jiang, M.
  2024{\natexlab{b}}.
\newblock Phys4DGen: A Physics-Driven Framework for Controllable and Efficient
  4D Content Generation from a Single Image.
\newblock \emph{arXiv preprint arXiv:2411.16800}.

\bibitem[{Lu et~al.()Lu, Wang, Sun, Chen, and Xie}]{luout}
Lu, W.; Wang, J.; Sun, X.; Chen, Y.; and Xie, X. ????
\newblock Out-of-distribution Representation Learning for Time Series
  Classification.
\newblock In \emph{The Eleventh International Conference on Learning
  Representations}.

\bibitem[{Majumdar et~al.(2022)Majumdar, Jadhav, Deodhar, Karande, Vig, and
  Runkana}]{PISN}
Majumdar, R.; Jadhav, V.; Deodhar, A.; Karande, S.; Vig, L.; and Runkana, V.
  2022.
\newblock Physics informed symbolic networks.
\newblock \emph{arXiv preprint arXiv:2207.06240}.

\bibitem[{Majumdar et~al.(2023)Majumdar, Jadhav, Deodhar, Karande, Vig, and
  Runkana}]{PINNSR2}
Majumdar, R.; Jadhav, V.; Deodhar, A.; Karande, S.; Vig, L.; and Runkana, V.
  2023.
\newblock Symbolic regression for pdes using pruned differentiable programs.
\newblock \emph{arXiv preprint arXiv:2303.07009}.

\bibitem[{Mundhenk et~al.(2021)Mundhenk, Landajuela, Glatt, Santiago, Petersen
  et~al.}]{SR2021M}
Mundhenk, T.; Landajuela, M.; Glatt, R.; Santiago, C.~P.; Petersen, B.~K.;
  et~al. 2021.
\newblock Symbolic regression via deep reinforcement learning enhanced genetic
  programming seeding.
\newblock \emph{Advances in Neural Information Processing Systems}, 34:
  24912--24923.

\bibitem[{Oh et~al.(2023)Oh, Amici, Bomarito, Zhe, Kirby, and
  Hochhalter}]{GPSR}
Oh, H.; Amici, R.; Bomarito, G.; Zhe, S.; Kirby, R.; and Hochhalter, J. 2023.
\newblock Genetic programming based symbolic regression for analytical
  solutions to differential equations.
\newblock \emph{arXiv preprint arXiv:2302.03175}.

\bibitem[{Petersen et~al.(2019)Petersen, Landajuela, Mundhenk, Santiago, Kim,
  and Kim}]{SR2021P}
Petersen, B.~K.; Landajuela, M.; Mundhenk, T.~N.; Santiago, C.~P.; Kim, S.~K.;
  and Kim, J.~T. 2019.
\newblock Deep symbolic regression: Recovering mathematical expressions from
  data via risk-seeking policy gradients.
\newblock \emph{arXiv preprint arXiv:1912.04871}.

\bibitem[{Polycarpou(2022)}]{FEM2022P}
Polycarpou, A.~C. 2022.
\newblock \emph{Introduction to the finite element method in electromagnetics}.
\newblock Springer Nature.

\bibitem[{Qin et~al.(2017)Qin, Song, Cheng, Cheng, Jiang, and
  Cottrell}]{qin2017dual}
Qin, Y.; Song, D.; Cheng, H.; Cheng, W.; Jiang, G.; and Cottrell, G.~W. 2017.
\newblock A dual-stage attention-based recurrent neural network for time series
  prediction.
\newblock In \emph{Proceedings of the 26th International Joint Conference on
  Artificial Intelligence}, 2627--2633.

\bibitem[{Raissi, Perdikaris, and Karniadakis(2019)}]{PINN}
Raissi, M.; Perdikaris, P.; and Karniadakis, G.~E. 2019.
\newblock Physics-informed neural networks: A deep learning framework for
  solving forward and inverse problems involving nonlinear partial differential
  equations.
\newblock \emph{Journal of Computational physics}, 378: 686--707.

\bibitem[{Robertsson and Blanch(2011)}]{FDM2020}
Robertsson, J.~O.; and Blanch, J.~O. 2011.
\newblock Numerical methods, finite difference.
\newblock In \emph{Encyclopedia of solid earth geophysics}, 883--892. Springer.

\bibitem[{Szab{\'o} and Babu{\v{s}}ka(2021)}]{FEM2021S}
Szab{\'o}, B.; and Babu{\v{s}}ka, I. 2021.
\newblock Finite element analysis: Method, verification and validation.

\bibitem[{Tenachi, Ibata, and Diakogiannis(2023)}]{tenachi2023deep}
Tenachi, W.; Ibata, R.; and Diakogiannis, F.~I. 2023.
\newblock Deep symbolic regression for physics guided by units constraints:
  toward the automated discovery of physical laws.
\newblock \emph{The Astrophysical Journal}, 959(2): 99.

\bibitem[{Udrescu and Tegmark(2020)}]{SR2020U}
Udrescu, S.-M.; and Tegmark, M. 2020.
\newblock AI Feynman: A physics-inspired method for symbolic regression.
\newblock \emph{Science advances}, 6(16): eaay2631.

\bibitem[{Ullah et~al.(2021)Ullah, Hayat, Ahmad, and Alhodaly}]{FDM2021U}
Ullah, H.; Hayat, T.; Ahmad, S.; and Alhodaly, M.~S. 2021.
\newblock Entropy generation and heat transfer analysis in power-law fluid
  flow: Finite difference method.
\newblock \emph{International Communications in Heat and Mass Transfer}, 122:
  105111.

\bibitem[{Wahyudi, Lestari, and Gapsari(2021)}]{FDM2021}
Wahyudi, S.; Lestari, P.; and Gapsari, F. 2021.
\newblock Application of Finite Difference Methods (FDM) on mathematical model
  of bioheat transfer of one-dimensional in human skin exposed environment
  condition.
\newblock \emph{J Mech Eng Res Develop}, 44(5): 1--9.

\bibitem[{Wang, Teng, and Perdikaris(2021)}]{GalerkinPINN}
Wang, S.; Teng, Y.; and Perdikaris, P. 2021.
\newblock Understanding and mitigating gradient flow pathologies in
  physics-informed neural networks.
\newblock \emph{SIAM Journal on Scientific Computing}, 43(5): A3055--A3081.

\bibitem[{Wen et~al.(2020)Wen, Sun, Yang, Song, Gao, Wang, and
  Xu}]{wen2020time}
Wen, Q.; Sun, L.; Yang, F.; Song, X.; Gao, J.; Wang, X.; and Xu, H. 2020.
\newblock Time series data augmentation for deep learning: A survey.
\newblock \emph{arXiv preprint arXiv:2002.12478}.

\bibitem[{Xu, Chen, and Wang(2024{\natexlab{a}})}]{xu2024kan4drift}
Xu, K.; Chen, L.; and Wang, S. 2024{\natexlab{a}}.
\newblock Kan4drift: Are kan effective for identifying and tracking concept
  drift in time series?
\newblock In \emph{NeurIPS Workshop on Time Series in the Age of Large Models}.

\bibitem[{Xu, Chen, and Wang(2024{\natexlab{b}})}]{xu2024kolmogorov}
Xu, K.; Chen, L.; and Wang, S. 2024{\natexlab{b}}.
\newblock Kolmogorov-arnold networks for time series: Bridging predictive power
  and interpretability.
\newblock \emph{arXiv preprint arXiv:2406.02496}.

\bibitem[{Yu et~al.(2018)}]{DeepRITZ}
Yu, B.; et~al. 2018.
\newblock The deep Ritz method: a deep learning-based numerical algorithm for
  solving variational problems.
\newblock \emph{Communications in Mathematics and Statistics}, 6(1): 1--12.

\bibitem[{Zeng, Gong, and Jiang(2023)}]{7}
Zeng, Q.; Gong, Y.; and Jiang, M. 2023.
\newblock Cross-Task Attack: A Self-Supervision Generative Framework Based on
  Attention Shift.
\newblock In \emph{IJCNN, 2023}. IEEE.

\bibitem[{Zhang et~al.(2019)Zhang, Song, Chen, Feng, Lumezanu, Cheng, Ni, Zong,
  Chen, and Chawla}]{zhang2019deep}
Zhang, C.; Song, D.; Chen, Y.; Feng, X.; Lumezanu, C.; Cheng, W.; Ni, J.; Zong,
  B.; Chen, H.; and Chawla, N.~V. 2019.
\newblock A deep neural network for unsupervised anomaly detection and
  diagnosis in multivariate time series data.
\newblock In \emph{Proceedings of the AAAI conference on artificial
  intelligence}, volume~33, 1409--1416.

\bibitem[{Zhang et~al.(2025)Zhang, Chen, Banzhaf, Zhang et~al.}]{zhang2025rag}
Zhang, H.; Chen, Q.; Banzhaf, W.; Zhang, M.; et~al. 2025.
\newblock RAG-SR: Retrieval-augmented generation for neural symbolic
  regression.
\newblock In \emph{The Thirteenth International Conference on Learning
  Representations}.

\bibitem[{Zhou et~al.(2022)Zhou, Ma, Wen, Wang, Sun, and
  Jin}]{zhou2022fedformer}
Zhou, T.; Ma, Z.; Wen, Q.; Wang, X.; Sun, L.; and Jin, R. 2022.
\newblock Fedformer: Frequency enhanced decomposed transformer for long-term
  series forecasting.
\newblock In \emph{International conference on machine learning}, 27268--27286.
  PMLR.

\end{thebibliography}

\end{document}